\documentclass[a4paper, 10pt, conference]{ieeeconf}      %
\usepackage{FG2026}

\FGfinalcopy %

\definecolor{iccvblue}{rgb}{0.21,0.49,0.74}
\usepackage[pagebackref,breaklinks,colorlinks,allcolors=iccvblue]{hyperref}

\usepackage{tcolorbox}
\usepackage{verbatim}
\usepackage{amsmath}
\usepackage{colortbl} 
\usepackage{graphicx}
\usepackage{multirow}  
\usepackage{pifont}
\usepackage{tikz}
\usepackage{cancel}
\usepackage{rotating}
\usepackage[dvipsnames]{xcolor}
\usepackage{lineno}
\usepackage{booktabs}
\usepackage{xspace}
\usepackage{cleveref}
\usepackage{amssymb}
\usepackage{listings}

\let\titleold\title
\renewcommand{\title}[1]{\titleold{#1}\newcommand{\thetitle}{#1}}
\def\maketitlesupplementary
{
\newpage
   \twocolumn[
    \centering
    \Large
    \textbf{\thetitle}\\
    \vspace{0.5em}Supplementary Material \\
    \vspace{1.0em}
   ] %
}

\newcommand{\ie}{\textit{i.e.}\xspace}
\newcommand{\eg}{\textit{e.g.}\xspace}

\lstdefinestyle{prompt}{
  basicstyle=\ttfamily\fontsize{7}{7.5}\selectfont,
  columns=fullflexible,
  breaklines=true,
  breakatwhitespace=false,
  breakindent=0em,
  keepspaces=true,
  showstringspaces=false,
  backgroundcolor=\color{ShadedGray},
  tabsize=2,
  captionpos=b,
}
\definecolor{ShadedGray}{RGB}{238,238,238}
\definecolor{ModelLightBlue}{RGB}{209, 233, 246}
\definecolor{DrawioBlue}{RGB}{218,232,252}
\definecolor{OurMethodColor}{RGB}{221, 241, 244}
\definecolor{DrawioOrange}{RGB}{255,230,204}
\definecolor{DrawioGreen1}{RGB}{204,255,204}
\definecolor{DrawioGreen}{RGB}{213,232,212}
\definecolor{DrawioPurple}{RGB}{225,213,231}
\definecolor{DrawioRed}{RGB}{248,206,204}
\definecolor{DrawioYellow}{RGB}{255, 242, 207}
\definecolor{DrawioBrightBlue}{RGB}{192,213,255}
\definecolor{DrawioBrightOrange}{RGB}{243,199,141}
\definecolor{DrawioBrightPurple}{RGB}{192,134,216}
\definecolor{MoreVividModelGreen}{RGB}{130, 200, 120}

\definecolor{R1color}{RGB}{244, 158, 76}
\definecolor{R2color}{RGB}{0, 175, 84}

\newcommand{\methodnameFull}{Training-Free Semantic Multi-Object Tracking with Vision--Language Models\xspace}

\newcommand{\methodname}{\textsc{TF-SMOT}\xspace}

\newcommand{\dataset}{BenSMOT\xspace}

\definecolor{LineBlue}{RGB}{045, 114, 255}

\definecolor{cvprblue}{rgb}{0.21,0.49,0.74}

\IEEEoverridecommandlockouts                              %
\def\FGPaperID{XXX} %

\title{\LARGE \bf
Training-Free Semantic Multi-Object\\Tracking with Vision-Language Models
}

\author{\parbox{16cm}{\centering
    {\large Laurence Bonat$^{1}$, Francesco Tonini$^{1, 2}$, Elisa Ricci$^{1,2}$, and Lorenzo Vaquero$^2$}\\
    {\normalsize
    $^1$Department of Information Engineering and Computer Science, University of Trento, Trento, Italy\\
    $^2$Fondazione Bruno Kessler, Trento, Italy}}
}

\usepackage{fancyhdr}
\begin{document}

\ifFGfinal
\thispagestyle{empty}
\pagestyle{empty}
\else
\author{Anonymous FG2026 submission\\ Paper ID 388 \\}
\pagestyle{plain}
\fi
\maketitle
\thispagestyle{fancy}
\renewcommand{\headrulewidth}{0pt}
\fancyhf{}
\fancyhead[C]{2026 International Conference on Automatic Face and Gesture Recognition (FG)}

\begin{abstract}
Semantic Multi-Object Tracking (SMOT) extends multi-object tracking with semantic outputs such as video summaries, instance-level captions, and interaction labels, aiming to move from trajectories to human-interpretable descriptions of dynamic scenes. 
Existing SMOT systems are trained end-to-end, coupling progress to expensive supervision, limiting the ability to rapidly adapt to new foundation models and new interactions.
We propose \methodname, a training-free SMOT pipeline that composes pretrained components for detection, mask-based tracking, and video-language generation.
\methodname combines D-FINE and the promptable SAM2 segmentation tracker to produce temporally consistent tracklets, uses contour grounding to generate video summaries and instance captions with InternVideo2.5, and aligns extracted interaction predicates to \dataset WordNet synsets via gloss-based semantic retrieval with LLM disambiguation.
On \dataset, \methodname achieves state-of-the-art \emph{tracking} performance within the SMOT setting and improves summary and caption quality compared to prior art.
Interaction recognition, however, remains challenging under strict exact-match evaluation on the fine-grained and long-tailed WordNet label space; our analysis and ablations indicate that semantic overlap and label granularity substantially affect measured performance.
\end{abstract}

\section{Introduction}

Multi-Object Tracking (MOT) is a core building block for video understanding, enabling applications ranging from autonomous systems to sports analytics and surveillance. 
However, MOT outputs are typically limited to trajectories and identities, leaving open the question of how to convert low-level tracks into high-level, human-interpretable descriptions of \emph{what} tracked entities do and \emph{how} they interact.
Recent work has therefore moved toward \emph{Semantic MOT} (SMOT), which couples tracking with language generation and structured interaction labels, targeting outputs such as video summaries, instance-level captions, and relational predicates~\cite{liMOTSemanticMultiObject2024}.
\emph{Success} in SMOT is inherently multi-objective: a system must (i) maintain temporally consistent identities and localization and (ii) produce semantic outputs that are faithful to the tracked evidence.
In practice, these semantic subtasks differ in difficulty: summaries and instance captions can tolerate lexical variability, while interaction labels require discrete, role-aware decisions (\ie, who does what to whom).

\begin{figure}[t]
    \centering
    \includegraphics[width=\linewidth]{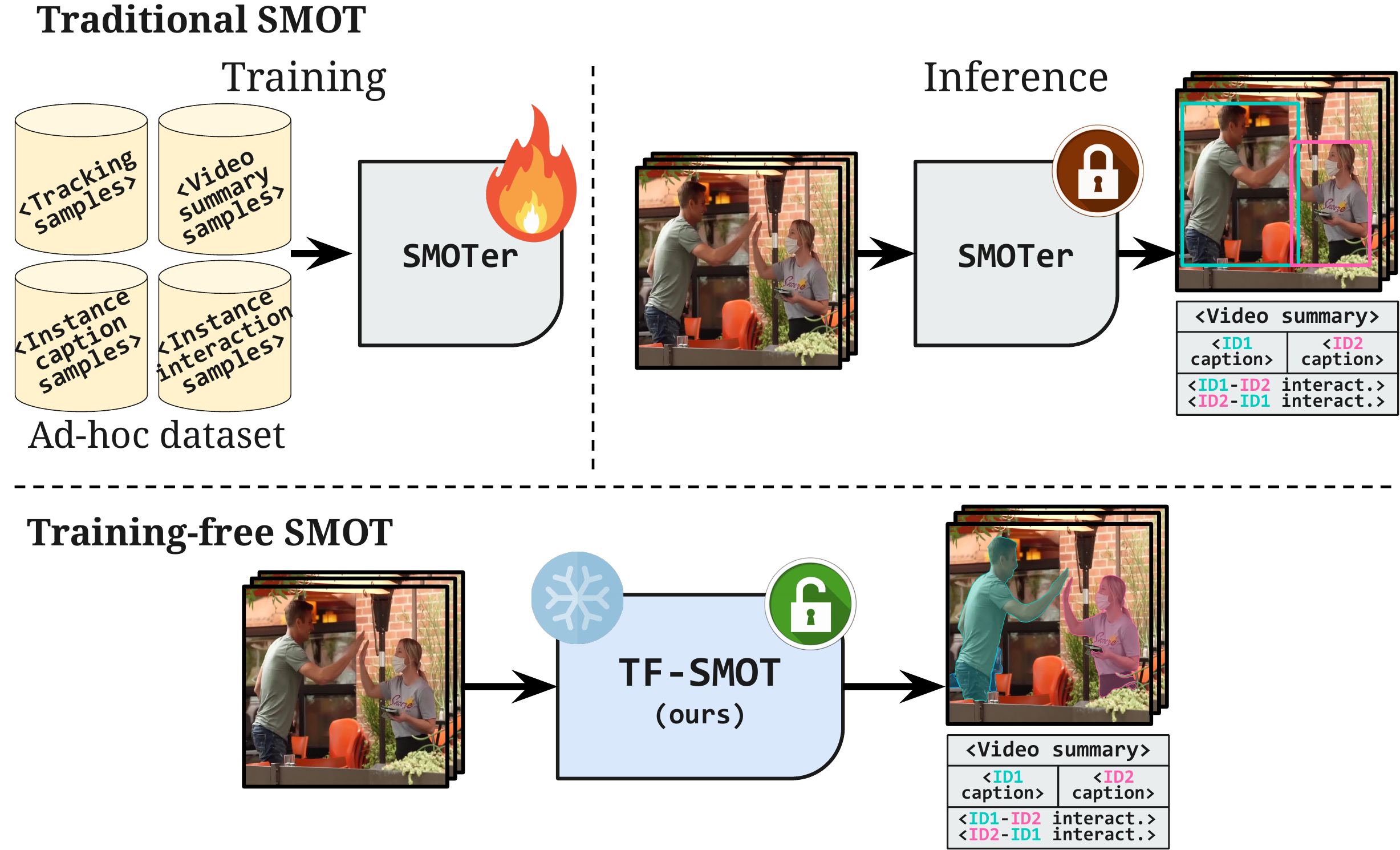}
    \caption{\textbf{Comparison of training-based and training-free Semantic Multi-Object Tracking (SMOT) paradigms.} Top: Traditional SMOT systems like SMOTer~\cite{liMOTSemanticMultiObject2024} require task-specific end-to-end training on annotated datasets, coupling tracking and semantic prediction to supervised learning. Bottom: our TF-SMOT operates entirely at inference time by composing frozen, pretrained components to produce tracking outputs together with video summaries, instance-level captions, and directed interactions, without using SMOT training data.}
    \label{fig:teaser}
\end{figure}

Despite progress, most SMOT systems are trained end-to-end and require task-specific architectures and supervision~\cite{liMOTSemanticMultiObject2024} (Fig.~\ref{fig:teaser}, top). 
This limits reproducibility, increases computational cost, and slows the adoption of rapidly improving foundation models.
At the same time, general-purpose models for detection, segmentation, video captioning, and language reasoning have recently reached a level of maturity that makes a training-free alternative plausible: modern detectors such as D-FINE~\cite{pengDFINERedefineRegression2024} improve localization, promptable video segmentation models such as SAM2~\cite{raviSAM2Segment2024} provide robust instance tracking via masks, and video-language models such as InternVideo2.5~\cite{wangInternVideo25EmpoweringVideo2025} can generate detailed long-context descriptions.

We propose \textbf{\methodname} (\methodnameFull), a training-free SMOT pipeline that decomposes SMOT into independently upgradable primitives and composes pretrained components to produce tracklets together with semantic outputs.
Central to our approach is \emph{contour-based instance grounding}: instead of cropping, boxing, or overlaying opaque masks, we convert segmentation masks into thin contours (Fig.~\ref{fig:teaser}, bottom) that act as a minimally invasive grounding signal for video-language models.
Our hypothesis is that contours provide persistent identity focus while largely preserving the underlying appearance features, reducing occlusion and color biases introduced by full-mask overlays.
As a result, visual language models (VLMs) can attend more effectively to individual instances and generate more accurate descriptions.
Finally, we address interaction recognition by extracting directed interaction predicates from captions with a large language model (LLM), then aligning them to the \dataset label space via semantic retrieval over WordNet~\cite{millerWordNetLexicalDatabase1995} glosses.

\methodname is designed to be a strong, reproducible baseline that is easy to update as better foundation models become available.
This modularity provides practical advantages (\eg, no task-specific training, interchangeable components, and simplified maintenance), but it may trade off against task-specific solutions that are trained end-to-end for a fixed benchmark.
In particular, our study reveals that interaction recognition on the \dataset benchmark~\cite{liMOTSemanticMultiObject2024} is sensitive to label granularity, semantic overlap, and non-exhaustive annotations, motivating improved benchmarks and evaluation protocols for SMOT.
Nonetheless, \methodname achieves state-of-the-art \emph{tracking} performance within the SMOT setting and improves language generation for summaries and instance captions over SMOTer~\cite{liMOTSemanticMultiObject2024}.
The main contributions of this work are:
\begin{itemize}
    \item We introduce \methodname, a \emph{training-free} SMOT pipeline that decomposes the task into modular primitives (person tracking, instance captioning, and interaction extraction) and serves as a strong training-free baseline.
    \item We propose contour-based instance grounding, a minimally invasive visual prompt that improves instance-centric captioning by focusing generation on a tracked identity while preserving appearance cues.
    \item We conduct an in-depth analysis of interaction recognition, showing how long-tailed label distributions and fine-grained WordNet-based taxonomies challenge current foundation models, and how coarser or clustered label spaces alleviate these issues.
    \item We demonstrate that \methodname achieves state-of-the-art tracking performance on BenSMOT, and improves over prior work on video summarization and instance captioning, despite using no task-specific training.
\end{itemize}

The remainder of the paper is organized as follows. Section~\ref{sec:related} reviews related work on MOT, semantic tracking, and vision-language models. Section~\ref{sec:task} formally introduces the task of SMOT, and Section~\ref{sec:methodParent} presents our proposed training-free pipeline. Section~\ref{sec:experiments} details the experimental setup and results, and Section~\ref{sec:conclusion} presents the conclusions and future work.

\section{Related work}\label{sec:related}

\subsection{Multi-object tracking}

Traditional multi-object tracking (MOT) decomposes the task into object detection and temporal association~\cite{Dai2022}.
Early approaches relied on hand-crafted features, background subtraction, optical flow, and Kalman filtering, combined with combinatorial data association~\cite{yilmazObjectTrackingSurvey2006,VaqueroMB20,kalmanNewApproachLinear1960}.
With the advent of deep learning, tracking-by-detection pipelines built on top of strong detectors such as Faster R-CNN~\cite{renFasterRCNNRealTime2017} or YOLO~\cite{redmonYouOnlyLook2016} became dominant.
SORT~\cite{bewleySimpleOnlineRealtime2016} and DeepSORT~\cite{wojkeSimpleOnlineRealtime2017} popularized simple yet effective online trackers that combine per-frame detections with motion and appearance embeddings.
Subsequent works improved robustness through better motion modeling and association strategies~\cite{zhangByteTrackMultiobjectTracking2022,VaqueroXABM24,caoObservationCentricSORTRethinking2023}, and graph-based methods such as GNN3DMOT~\cite{wengGNN3DMOTGraphNeural2020}.

Transformer-based architectures have recently been adopted for MOT, often building on DETR-style object queries.
TrackFormer~\cite{meinhardtTrackFormerMultiObjectTracking2022}, TransTrack~\cite{sunTransTrackMultipleObjectTracking2020}, MOTR~\cite{zengMOTREndtoEndMultipleObject2022}, and MOTRv2~\cite{zhangMOTRv2BootstrappingEndtoEnd2023} reformulate tracking as a sequence prediction problem, leveraging attention to propagate object queries over time.
While these models achieve competitive performance, they typically require extensive task-specific training on large-scale tracking datasets.
Our work is complementary: we adopt an existing, training-free segmentation tracker, SAM2~\cite{raviSAM2Segment2024}, and focus instead on augmenting trajectories with semantic information using foundation models, without any additional supervision.

\subsection{Semantic tracking and human-human interactions}

Moving beyond geometric tracking, several works enrich trajectories with semantics.
Semantic MOT (SMOT) aims to attach natural-language descriptions and interaction labels to tracks, enabling higher-level understanding of multi-person scenes.
Li \emph{et al.}~\cite{liMOTSemanticMultiObject2024} introduced the BenSMOT benchmark, which provides human trajectories together with video-level summaries, instance captions, and WordNet-based interaction labels, and proposed SMOTer, an end-to-end model jointly trained for tracking and these semantic tasks.
Related research in human-object interaction (HOI) detection~\cite{hanSurveyHumanObjectInteraction2025,gaoICANInstanceCentricAttention2018,tonini2025dynamic,houVisualCompositionalLearning2020} tackles triplet prediction in images, but is usually not coupled with full multi-object tracking.

Our work shares the SMOT goal of producing both trajectories and rich semantics, but differs in two key aspects.
First, we explicitly target a training-free setting, relying solely on off-the-shelf models rather than training new task-specific architectures.
Second, instead of predicting interactions directly from visual features, we derive them from instance-level captions produced by a video vision-language model, and map extracted verbs to a lexical ontology (WordNet~\cite{millerWordNetLexicalDatabase1995}) with the help of an LLM.
This design exposes both the strengths and limitations of current foundation models for fine-grained interaction understanding.

\subsection{Vision-language and multimodal large language models}

Recent years have seen rapid progress in vision-language models (VLMs) and multimodal large language models (MLLMs).
Contrastive models such as CLIP~\cite{radford2021learning} learn joint image-text embeddings and enable zero-shot classification and retrieval.
These ideas have been extended to video with models such as VideoCLIP~\cite{xu2021videoclip} and InternVideo~\cite{wang2022internvideo}.
More recent MLLMs, including Flamingo~\cite{alayracFlamingoVisualLanguage2022}, BLIP-2~\cite{liBLIP2BootstrappingLanguageImage2023}, MiniGPT-4~\cite{zhuMiniGPT4EnhancingVisionLanguage2023}, LLaVA~\cite{liuImprovedBaselinesVisual2024}, and InternVideo2.5~\cite{wangInternVideo25EmpoweringVideo2025}, combine powerful language models with visual encoders to support open-ended multimodal reasoning and generation.

These models have been applied to a variety of video understanding tasks, such as dense captioning, question answering, and open-vocabulary recognition~\cite{zhouDenseVideoObject2024,abdarReviewDeepLearning2023,bordesIntroductionVisionLanguageModeling2024}.
However, their use in structured tracking scenarios remains relatively unexplored.
Closest to our setting, Type-to-Track~\cite{phanguyenTypetoTrackRetrieveAny2023} and related works use language prompts to select objects to track, while OW-VISCapTor~\cite{choudhuriOWVISCapTorAbstractorsOpenWorld2024} jointly performs open-world video instance segmentation and captioning.

\begin{figure*}[!ht]
    \centering
    \includegraphics[width=0.9\linewidth]{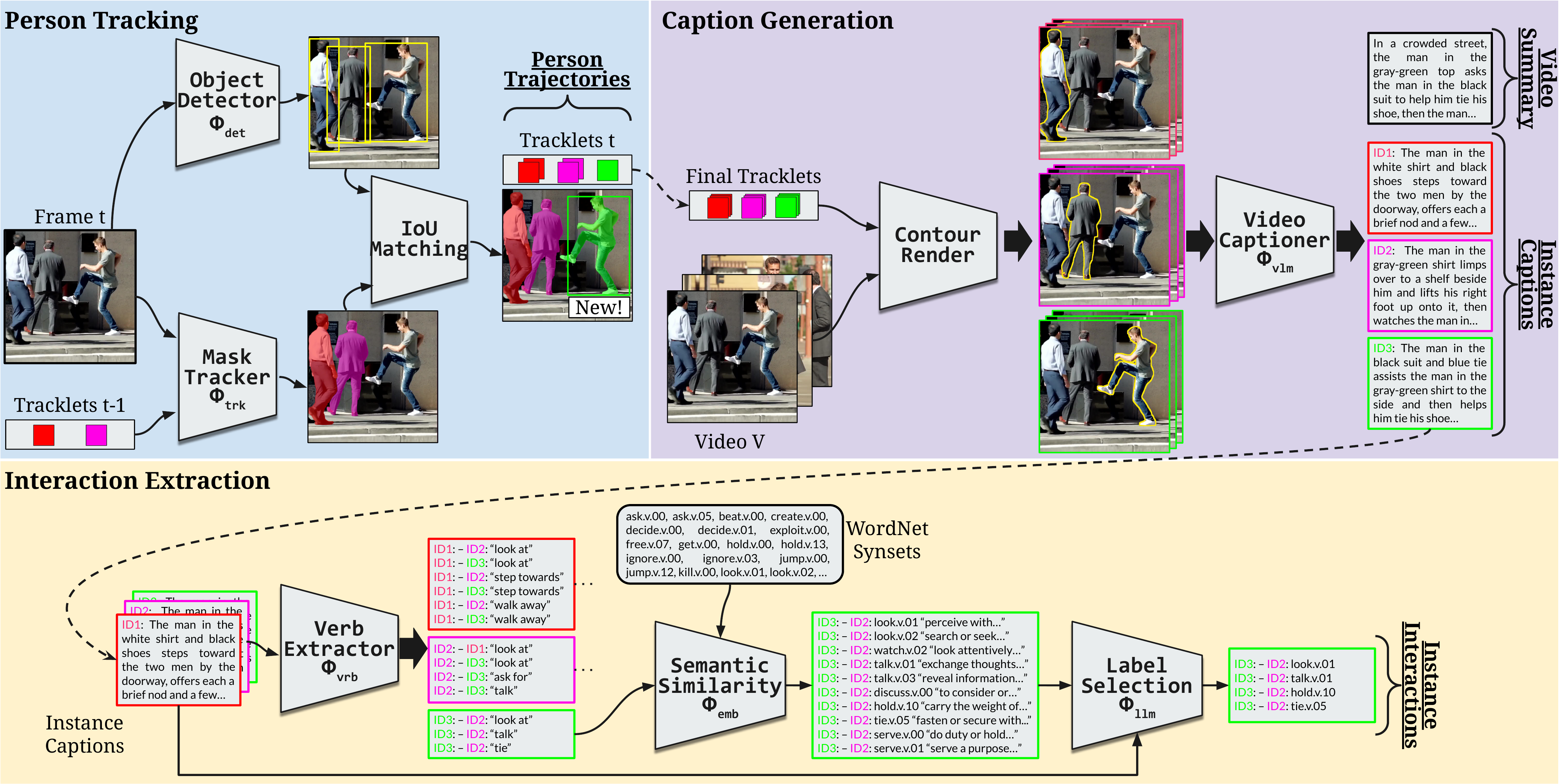}
    \caption{\textbf{\methodname overview.} The system decomposes SMOT into explicit functional stages with well-defined interfaces.
    The \emph{Person Tracking} module (left) utilizes a person detector $\Phi_{\text{det}}$ and mask-based tracker $\Phi_{\text{trk}}$ to determine the \emph{person trajectories} within the video.
    Subsequently, the \emph{Caption Generation} module (right) renders the contours of the people to better ground them and uses a video captioner $\Phi_{\text{vlm}}$ to generate the \emph{instance captions} and \emph{video summary}.
    Finally, the \emph{Interaction Extraction} module (bottom) identifies interaction verbs via $\Phi_{\text{vrb}}$, maps them to WordNet synsets using semantic similarity $\Phi_{\text{emb}}$, and selects the most relevant \emph{instance interactions} using an LLM $\Phi_{\text{emb}}$ with caption context.}
    \label{fig:pipeline}
    \vspace{-1em}
\end{figure*}

In parallel, large language models have been leveraged for structured information extraction and scene graph parsing~\cite{liFACTUALBenchmarkFaithful2023}, often aided by grammar-constrained decoding~\cite{dong2024xgrammar}.
We build on these advances by using InternVideo2.5 for video and instance captioning, and a text-only LLaMA~\cite{llamateam2024} combined with grammar control to extract directed human-human interactions and align them with WordNet-based labels.
Unlike most prior works, we deliberately refrain from any fine-tuning, and instead investigate how far these components can be pushed in a purely zero-shot, training-free configuration.

\section{Task Definition}\label{sec:task}
Let a video be a sequence of frames $\mathcal{V} = \{I_t\}_{t=1}^{T}$. 
SMOT requires (a) tracking all target instances through time and (b) producing semantic outputs grounded in these tracks. 
Following \dataset~\cite{liMOTSemanticMultiObject2024}, we consider four different types of outputs, as follows:
\begin{itemize}
    \item \textbf{Tracking trajectories} $\{b_{i,t}\}_{i=1,t=1}^{N,T}$, consisting of bounding box sequences $b_{i,t} = (x_{i,t}, y_{i,t}, w_{i,t}, h_{i,t})$ for each tracked instance $i$ across all $T$ frames, enabling spatio-temporal localization of targets and answering ``where are the targets?''.
    
    \item \textbf{Instance captions} $\{c_i\}_{i=1}^{N}$, where each $c_i$ provides a precise natural language description of person $i$'s behavior over its full trajectory, capturing fine-grained actions and attributes (e.g., ``A boy wearing black short-sleeves stretched out his hands and chased a boy wearing blue long-sleeves'').
    
    \item \textbf{Interaction triplets} capturing pairwise relations between tracked instances, expressed as WordNet~\cite{millerWordNetLexicalDatabase1995} synsets from a given label set $\mathcal{W}$, modeling ``who interacts with whom and how''.
    
    \item \textbf{Video summary} $s$ describing the video at a global level, integrating all trajectories, interactions, and scene context (e.g., ``On a small outdoor basketball court, three boys were playing basketball'').
\end{itemize}
In the following, we show how foundational multimodal models are a strong baseline for SMOT without any task-specific finetuning.

\section{Method}
\label{sec:methodParent}
Our proposed \methodname is a modular, training-free pipeline that composes pretrained components to produce tracks and grounded semantic outputs (Fig.~\ref{fig:pipeline}).
We first detect people and track each identity as a temporally consistent mask sequence (Sec.~\ref{subsec:tracking}).
Next, we render a contour-grounded clip for each identity and query a video VLM to generate a global video summary and per-instance captions (Sec.~\ref{subsec:captioning}).
Finally, we extract directed interaction predicates from captions and align them to \dataset's WordNet synset vocabulary (Sec.~\ref{subsec:interactions}).
All stages use off-the-shelf models and require no \dataset training data nor any additional finetuning.

\subsection{Person Tracking}\label{subsec:tracking}
As shown in Fig.~\ref{fig:pipeline}, left, we follow a tracking-by-detection strategy: a person detector proposes boxes that initialize a promptable segmentation tracker, which outputs per-frame masks for each identity.

\noindent\textbf{Detection.}
For each frame $I_t$, we apply an off-the-shelf person detector $\Phi_{\text{det}}$ to obtain a set of candidate boxes $\mathcal{B}_t=\{b_{t,k}\} = \Phi_{\text{det}}(I_t)$, filtering to the \texttt{person} class and discarding low-confidence detections.

\noindent\textbf{Mask tracking.}
We use a promptable segmentation tracker $\Phi_{\text{trk}}$ to track all detected people as instance masks.
Given an initialization prompt (we use the detector boxes $\{b_{t,i}\}$), the tracker returns a segmentation mask $m_{t,i}$ for each active identity $i$ at time $t$.
For management of identities in an online setting, we convert each mask into its tight bounding box $\Box(m_{t,i})$ and compare it with detector boxes in the current frame using intersection-over-union (IoU):
\begin{equation}
\mathrm{IoU}(b, b') = \frac{|b\cap b'|}{|b\cup b'|}.
\end{equation}
We treat a detection $b_{t,k}$ as a new identity if $\max_i \mathrm{IoU}(b_{t,k}, \Box(m_{t,i})) < \tau_{\mathrm{new}}$, and initialize the tracker with this box.
Importantly, this IoU test is used only for \emph{track initialization}: once an identity is created, per-frame association is handled internally by the tracker's memory mechanism rather than by greedy box matching.
Otherwise, the tracker updates the existing identity through its internal memory mechanism, which maintains temporal consistency and handles disappearance without explicit post-processing.

\subsection{Caption Generation}\label{subsec:captioning}
To appropriately describe individual instances, we first render thin mask contours as a minimally invasive visual prompt for instance-centric captioning (Fig.~\ref{fig:pipeline}, right), and then apply a vision–language model (VLM) to perform video-to-text generation.

\noindent\textbf{Contour rendering for grounding.}
To ground a video-language model (VLM) on a specific instance without corrupting appearance cues, we render only the contour of the target mask on top of each frame.
For each identity $i$ and frame $t$, we form a contour-grounded frame $\tilde{I}_{t,i}$ by overlaying the contour of $m_{t,i}$ on $I_t$, and denote the resulting clip as $\tilde{\mathcal{V}}_i=\{\tilde{I}_{t,i}\}_{t=1}^{T}$.
Using full opaque masks can bias the VLM toward mask color and occlude visual details; contour overlays provide persistent instance focus while preserving appearance information.
Compared to boxes, contours provide a tighter instance signal that follows articulation and scale changes; compared to crops, they preserve surrounding context needed to interpret interactions and role assignments.
We view contour grounding as an attention-guidance primitive: the contour acts as a low-bandwidth cue that nudges generation toward the intended identity while minimizing distribution shift in the underlying pixels.

\noindent\textbf{Instance Captions.}
We query a VLM $\Phi_{\text{vlm}}$ on each contour-grounded clip $\tilde{\mathcal{V}}_i$ and prompt it to generate a single, detailed sentence $c_i=\Phi_{\text{vlm}}(\tilde{\mathcal{V}}_i)$ focusing on the contoured person and its actions and interactions over time.
Because grounding is provided directly in the input pixels, contouring makes this stage model-agnostic and compatible with any video large language model used as a black box.
Finally, for the \textbf{video summary}, the model receives the raw video $\mathcal{V}$ and is prompted to output a single concise sentence $s=\Phi_{\text{vlm}}(\mathcal{V})$ describing the scene and the main human actions.

\subsection{Interaction Extraction}\label{subsec:interactions}
We infer directed interactions from the predicted instance captions in three, stages (Fig.~\ref{fig:pipeline}, bottom): we first extract free-form predicates describing ``who does what to whom'', then align each predicate to \dataset's WordNet synset label space $\mathcal{W}$, and finally select the most relevant labels.

\noindent\textbf{Predicate extraction.}
We use an LLM $\Phi_{\text{vrb}}$ to extract directed interaction predicates between tracked people from the caption set $\{c_i\}$.
The output is a collection of predicate sets $\{\mathcal{P}_{i\rightarrow j}\}_{i\neq j}$, where each $\mathcal{P}_{i\rightarrow j}$ contains zero or more short free-text predicates describing interactions from subject $i$ to object $j$.
To make outputs reliably machine-readable, we constrain decoding to a JSON schema using xGrammar~\cite{dong2024xgrammar} and request base-form predicates, which simplifies downstream normalization.
We focus on predicate-like labels because (i) they are easier to extract from captions than full scene graphs, and (ii) \dataset interaction annotations are expressed as WordNet synsets, where semantically adjacent labels are frequently differentiated by subtle sense definitions.

\noindent\textbf{Alignment to $\mathcal{W}$.}
\dataset labels interactions as WordNet synsets (e.g., \texttt{talk.v.01}) with associated natural-language glosses~\cite{millerWordNetLexicalDatabase1995}.
For each extracted predicate $p\in\mathcal{P}_{i\rightarrow j}$, we compute semantic similarity against all candidate synsets $w\in\mathcal{W}$ using a sentence embedding model $\Phi_{\text{emb}}$ applied to the predicate and to the synset gloss $g(w)$:
\begin{equation}
s(p,w) = \cos\left(\Phi_{\text{emb}}(p), \Phi_{\text{emb}}(g(w))\right),
\end{equation}
and keep the top-$K$ candidates by $s(p,w)$.

\noindent\textbf{Context-aware label selection.}
Cosine similarity alone often struggles to disambiguate fine-grained senses.
To incorporate context, we ask a large language model $\Phi_{\text{llm}}$ to select, for each predicate and subject-object pair, the most relevant interaction among the top-$K$ candidates.
The LLM is provided with (i) the original instance captions for the subjects, (ii) the predicate $p$, and (iii) the list of candidate synsets and their glosses, and is instructed to return only the index of the most appropriate synset.
This yields a predicted synset set for each ordered pair $\hat{\mathcal{R}}_{i\rightarrow j}=\{\hat{w}(p)\mid p\in\mathcal{P}_{i\rightarrow j}\}\subseteq \mathcal{W}$, which we compare against ground truth for evaluation.
All interaction extraction and alignment operate on the predicted instance captions $\{c_i\}$ (not ground-truth language).
This hybrid design leverages sentence embeddings for efficient semantic retrieval while delegating fine-grained disambiguation to the LLM, and can be easily adapted to different label spaces by changing the candidate synset set and glosses. 

\begin{table}[t]
\centering
\caption{Top 30 frequent interactions and their number of occurrences in the \dataset test set.}
\resizebox{0.99\columnwidth}{!}{%
\begin{tabular}{lrlrlr}
\toprule
    look.v.01 & 2353 & touch.v.01 & 261 & facial.n.02 & 130 \\
    smile.v.01 & 1291 & show.v.01 & 239 & rally.n.05 & 127 \\
    talk.v.01 & 1047 & comply.v.01 & 184 & hash\_out.v.01 & 108 \\
    talk.v.02 & 714 & receive.v.01 & 171 & insert.v.01 & 104 \\
    converse.v.01 & 663 & dance.v.01 & 166 & teach.v.01 & 98 \\
    listen.v.01 & 468 & handshake.n.01 & 166 & help.v.01 & 96 \\
    embrace.v.02 & 339 & play.v.01 & 165 & pass.v.20 & 92 \\
    hold.v.02 & 332 & baptism.n.01 & 148 & sing.v.01 & 88 \\
    collaborate.v.01 & 314 & box.v.02 & 148 & order.v.01 & 82 \\
    accept.v.02 & 274 & give.v.03 & 143 & propose.v.05 & 78 \\
\bottomrule
\end{tabular}
}%
\label{tab:interaction_stats}
\end{table}

\section{Experiments}
\label{sec:experiments}

\subsection{Dataset}
We evaluate on \dataset~\cite{liMOTSemanticMultiObject2024}, a human-centric SMOT benchmark containing 3292 videos across 47 scenarios.
Each video is annotated with (i) multi-person tracks, (ii) a video-level summary sentence, (iii) instance-level captions, and (iv) directed interactions between tracked people labeled as WordNet synsets.
Following the training-free setting of this work, we do not use the training split and report results on the official test split.

\noindent\textbf{Interaction label distribution.}
\dataset contains 335 interaction labels, 327 of which are WordNet synsets~\cite{millerWordNetLexicalDatabase1995}.
As shown in Table~\ref{tab:interaction_stats}, the distribution is strongly long-tailed: $75\%$ of interaction instances are concentrated within the top 30 labels, and the most frequent label \texttt{look.v.01} accounts for $16\%$ of all interactions.
This skew makes exact-match interaction evaluation sensitive to rare classes and semantically overlapping synsets.

\subsection{Metrics}
For tracking, we report Higher Order Tracking Accuracy (HOTA) and its decomposition into association (AssA) and detection (DetA), along with localization accuracy (LocA)~\cite{luitenHOTAHigherOrder2021}. 
We also report CLEAR MOTA~\cite{bernardinEvaluatingMultipleObject2008}, the number of identity switches (IDSW), and IDF1~\cite{ristaniPerformanceMeasuresData2016}.
For language generation (summaries and captions) we use BLEU~\cite{papineniBLEUMethodAutomatic2001}, METEOR~\cite{banerjeeMETEORAutomaticMetric2005}, ROUGE-L~\cite{linROUGEPackageAutomatic2004}, and CIDEr~\cite{vedantamCIDErConsensusbasedImage2015}.
For interaction recognition, we report precision, recall, and F1 over the WordNet synset labels.

\subsection{Implementation details}
\methodname uses D-FINE~\cite{pengDFINERedefineRegression2024} for person detection $\Phi_{\text{det}}$ and SAM2~\cite{raviSAM2Segment2024} for mask-based tracking $\Phi_{\text{trk}}$, initialized from detector boxes and updated online in a forward (causal) pass without accessing future frames.
We discard detections with confidence $p<0.8$ to avoid initializing spurious tracks.
For contour grounding (Sec.~\ref{subsec:captioning}), we render a $5$-pixel-wide contour around the target mask and generate one contour-grounded clip per identity (rendering only the target contour) to avoid distractions from other instances.
We generate video summaries and instance captions $\Phi_{\text{vlm}}$ with InternVideo2.5~\cite{wangInternVideo25EmpoweringVideo2025}, queried once per video for the summary and once per identity for instance captions using the full contour-grounded clip (no sliding windows).
For interactions the verb extraction $\Phi_{\text{vrb}}$ and label selection $\Phi_{\text{llm}}$, we use a LLaMA 3.1 8B~\cite{llamateam2024} with xGrammar-constrained decoding~\cite{dong2024xgrammar} to extract directed predicates, and align them to WordNet synsets in \dataset using all-MiniLM-L6-v2~\cite{WangW0B0020} as the sentence embedding model $\Phi_{\text{emb}}$ (top-$K$ with $K=5$).
All components are applied without task-specific training or fine-tuning.
Refer to~\cref{sec:supp_prompt} in Sup.Mat. for the prompts used in \methodname.

Training-based methods like SMOTer~\cite{liMOTSemanticMultiObject2024} are trained end-to-end supervisedly on \dataset, while our \methodname composes large pre-trained models (detectors, segmentation trackers, and VLMs/LLMs) that were trained on substantially larger, web-scale data.
Therefore, our comparisons should not be interpreted as a controlled ``same-data'' or ``same-supervision'' evaluation.
We position \methodname as a \emph{training-free} baseline that quantifies what can be achieved without access to \dataset training data, and as a tool to analyze which SMOT subtasks are well served by current foundation models (tracking and captioning) and which remain brittle (fine-grained interaction labels).
Accordingly, \methodname is not intended to replace trained SMOT systems, but to provide reference point for the research community.

\subsection{Comparison with State of the Art}

\begin{table*}[!t]
\centering
\caption{Tracking performance on \dataset. Best results are in \textbf{bold}}
\resizebox{0.9\linewidth}{!}{%
\begin{tabular}{l|ccccccccc}
    \toprule[1.2pt]
    \textbf{Method} & \textbf{HOTA}$\uparrow$ & \textbf{AssA}$\uparrow$ & \textbf{DetA}$\uparrow$ & \textbf{LocA}$\uparrow$ & \textbf{MOTA}$\uparrow$ & \textbf{IDSW}$\downarrow$ & \textbf{IDR}$\uparrow$ & \textbf{IDP}$\uparrow$ & \textbf{IDF1}$\uparrow$\\
    \midrule[0.8pt]
    SORT~\cite{bewleySimpleOnlineRealtime2016} & 48.49 & 38.95 & 60.91 & 87.50 & 53.58 & 13875 & 60.85 & 48.43 & 53.93 \\
    DeepSORT~\cite{wojkeSimpleOnlineRealtime2017} & 50.12 & 40.23 & 61.45 & 87.67 & 54.29 & 11278 & 62.10 & 51.11 & 56.76 \\
    OC-SORT~\cite{caoObservationCentricSORTRethinking2023} & 51.00 & 41.42 & 63.31 & 87.61 & 55.19 & 15049 & 63.92 & 53.10 & 58.01 \\
    ByteTrack~\cite{zhangByteTrackMultiobjectTracking2022} & 68.84 & 71.15 & 67.10 & 85.15 & 73.87 & 1712 & 82.25 & 74.83 & 78.37 \\
    TransTrack~\cite{sunTransTrackMultipleObjectTracking2020} & 71.31 & 73.34 & 69.67 & \textit{91.31} & 74.08 & 2530 & \textit{85.63} & 72.75 & 78.67 \\
    MOTR~\cite{zengMOTREndtoEndMultipleObject2022} & 66.10 & 73.12 & 55.14 & 86.30 & 45.19 & 617 & 72.39 & 70.12 & 68.97 \\
    MOTRv2~\cite{zhangMOTRv2BootstrappingEndtoEnd2023} & 65.28 & \textit{76.82} & 51.30 & 86.09 & 45.52 & \textit{430} & 78.47 & 65.51 & 70.76 \\
    SMOTer~\cite{liMOTSemanticMultiObject2024} & \textit{71.98} & 73.71 & \textit{70.79} & 87.11 & \textit{77.71} & 1702 & 83.82 & \textit{77.97} & \textit{80.65} \\
    \midrule[0.8pt]
    \textbf{\methodname (ours)} & \textbf{85.47} & \textbf{94.06} & \textbf{77.75} & \textbf{94.68} & \textbf{80.92} & \textbf{45} & \textbf{86.74} & \textbf{93.48} & \textbf{89.99} \\
    \bottomrule[1.2pt]
\end{tabular}%
}
\label{tab:tracking}
\end{table*}

\begin{table*}[!t]
\centering
\caption{Language generation results (video summary and instance captioning) on \dataset. Best results are in \textbf{bold}.}
\resizebox{0.9\linewidth}{!}{%
\begin{tabular}{l|cccc|cccc}
\toprule
        \multirow{2}{*}{\textbf{Method}} & \multicolumn{4}{c|}{\textbf{Video Summary}} & \multicolumn{4}{c}{\textbf{Instance Captioning}} \\
        & \textbf{BLEU} & \textbf{METEOR} & \textbf{ROUGE} & \textbf{CIDEr} & \textbf{BLEU} & \textbf{METEOR} & \textbf{ROUGE} & \textbf{CIDEr} \\
        \midrule
        SMOTer~\cite{liMOTSemanticMultiObject2024} & 0.245 & 0.261 & 0.223 & 0.343 & 0.306 & 0.209 & 0.223 & \textbf{0.087} \\
        \textbf{\methodname (ours)} & \textbf{0.369} & \textbf{0.262} & \textbf{0.281} & \textbf{0.360} & \textbf{0.383} & \textbf{0.239} & \textbf{0.228} & \textbf{0.087} \\
    \bottomrule[1.2pt]
\end{tabular}%
}
\label{tab:language}
\end{table*}

\noindent\textbf{Tracking.}
Table~\ref{tab:tracking} compares \methodname against standard tracking-by-detection baselines (SORT~\cite{bewleySimpleOnlineRealtime2016}, DeepSORT~\cite{wojkeSimpleOnlineRealtime2017}, OC-SORT~\cite{caoObservationCentricSORTRethinking2023}, ByteTrack~\cite{zhangByteTrackMultiobjectTracking2022}), transformer-based trackers (TransTrack~\cite{sunTransTrackMultipleObjectTracking2020}, MOTR~\cite{zengMOTREndtoEndMultipleObject2022}, MOTRv2~\cite{zhangMOTRv2BootstrappingEndtoEnd2023}), and the SMOTer baseline~\cite{liMOTSemanticMultiObject2024}.
\methodname substantially outperforms SMOTer across all tracking metrics, achieving +13.49 on HOTA (85.47 vs 71.98), +20.35 on AssA (94.06 vs 73.71), +6.96 on DetA (77.75 vs 70.79), +7.57 on LocA (94.68 vs 87.11), +3.21 on MOTA (80.92 vs 77.71), +9.34 on IDF1 (89.99 vs 80.65), -1657 on IDSW (45 vs 1702), +2.92 on IDR (86.74 vs 83.82), and +15.51 on IDP (93.48 vs 77.97).

\noindent\textbf{Instance and summary caption evaluation.}
Table~\ref{tab:language} reports summary and instance-caption metrics.
Contour-based grounding improves instance captioning compared to SMOTer, while a general-purpose LLM yields substantially better summaries than the fine-tuned one within SMOTer.
For the video summary, \methodname outperforms BenSMOT on all metrics, achieving +0.124 on BLEU (0.369 vs 0.245), +0.001 on METEOR (0.262 vs 0.261), +0.058 on ROUGE (0.281 vs 0.223), +0.017 on CIDEr (0.360 vs 0.343).
For instance captioning we see a similar trend: +0.077 on BLEU (0.383 vs 0.306), +0.030 on METEOR (0.239 vs 0.209), +0.005 on ROUGE (0.228 vs 0.223), comparable results for CIDEr.
We note that CIDEr for instance captions remains comparable, despite higher BLEU/METEOR; this is consistent with our captions being longer and more detailed, which can change n-gram weighting and penalize stylistic divergence even when semantics are correct.

\noindent\textbf{Exact-match interaction recognition.}
Table~\ref{tab:interactions} reports strict exact-match interaction recognition on the full WordNet synset space, which conflates semantic understanding with ontology alignment.
To contextualize this difficulty, we also report lemma-merged and coarse-cluster interaction scores for \methodname; these relaxed settings are not directly comparable to strict WordNet evaluation but highlight the impact of label granularity.
As shown, \methodname exhibits a lower F1 score than the fully-supervised SMOTer.
This occurs because, although \methodname produces relevant and coherent interaction predictions (Fig.~\ref{fig:qualitative}), many are not included in the ground truth, and exact-match evaluation penalizes semantically adjacent synsets.

For example, for one interaction pair the model predicts \{\texttt{converse.v.01}, \texttt{talk.v.01}\} while the ground truth contains only \{\texttt{talk.v.01}\}, illustrating near-miss ambiguity in the WordNet label space.
We observe three recurring sources of failure, summarized in Table~\ref{tab:int_taxonomy}:
\begin{itemize}
    \item \textbf{Sense granularity}: interactions are abstract and context-dependent, and multiple WordNet verb senses can be partially correct for the same event.
    \item \textbf{Retrieval bottlenecks}: gloss-based retrieval can mis-rank subtle senses because embedding spaces conflate semantically adjacent definitions; this is consistent with the top-1 cosine ablation strongly reducing recall (Tab.~\ref{tab:int_ablation}).
    \item \textbf{Role and directionality}: directed interactions require correct subject-object assignment, and role ambiguity in captions can flip the predicted direction even when the underlying event is understood.
\end{itemize}
Furthermore, the long-tailed distribution amplifies these effects: many synsets occur only a handful of times in \dataset, making rare interactions difficult to recover without supervision or a coarser label space.

\begin{table}[!t]
\centering
\caption{Strict exact-match interaction recognition on \dataset (top), and \methodname under relaxed label spaces (bottom).}
\resizebox{0.99\linewidth}{!}{%
\begin{tabular}{l|ccc}
\toprule[1.2pt]
\textbf{Method / evaluation} & \textbf{Prec.} & \textbf{Rec.} & \textbf{F1} \\
\midrule
SMOTer (WordNet synsets)~\cite{liMOTSemanticMultiObject2024} & 0.434 & 0.320 & 0.368 \\
\methodname (WordNet synsets) & 0.089 & 0.116 & 0.101 \\
\midrule
\methodname (lemma-merged synsets)$^\star$ & 0.080 & 0.103 & 0.090 \\
\methodname (20 coarse clusters)$^\star$ & 0.495 & 0.207 & 0.292 \\
\bottomrule[1.2pt]
\multicolumn{4}{l}{$^\star$Not directly comparable across settings.}
\end{tabular}%
}
\label{tab:interactions}
\end{table}

\begin{table}[!t]
\centering
\caption{Interaction failure taxonomy for WordNet-synset evaluation on \dataset.}
\resizebox{0.95\linewidth}{!}{%
    \begin{tabular}{@{}p{0.26\linewidth}p{0.70\linewidth}@{}}
        \toprule[1.2pt]
        \textbf{Error type} & \textbf{Typical manifestation under strict evaluation} \\
        \midrule
        Sense ambiguity & Multiple synsets can be plausible for the same event (e.g., \texttt{talk.v.01} vs.\ \texttt{converse.v.01}), yet not all of them are present in the ground truth labels. \\
        Near-synonym mismatch & Predictions fall in a tight semantic neighborhood but are counted fully wrong under exact match. \\
        Direction / role flip & Subject/object attribution swaps, flipping directed interactions even when the event is recognized. \\
    \bottomrule[1.2pt]
\end{tabular}%
}
\label{tab:int_taxonomy}
\end{table}

\subsection{Ablations on Interaction Prediction}
Given the drop in interaction F1 (Tab.~\ref{tab:interactions}), we report controlled ablations that isolate two factors: (i) the retrieval/disambiguation mechanism and (ii) the effective label space.
Table~\ref{tab:int_ablation} shows that removing LLM disambiguation (top-1 cosine) collapses recall, while keeping multiple candidates (top-5 cosine) substantially increases recall but at low precision.
Reducing redundancy by merging synsets that share the same lemma slightly improves F1, suggesting that label noise and semantic overlap affect evaluation.
Finally, restricting the label space to frequent interactions or clustering into 20 coarse categories yields much higher F1, highlighting the tension between fine-grained labels and reliable prediction under a training-free setup.

\begin{table}[!t]
\centering
\caption{Interaction ablations.}
\resizebox{0.99\linewidth}{!}{%
    \begin{tabular}{l|l|ccc}
        \toprule[1.2pt]
        \textbf{Label space / setting} & \textbf{Selector} & \textbf{Prec.} & \textbf{Rec.} & \textbf{F1} \\
        \midrule
        \multirow{3}{*}{Full WordNet (335)} & LLM (ours) & 0.089 & 0.116 & 0.101 \\
         & top-1 cosine & 0.032 & 0.043 & 0.037 \\
         & top-5 cosine & 0.039 & 0.239 & 0.067 \\
        \midrule
        \multirow{3}{*}{Lemma-merged (259)} & LLM & 0.080 & 0.103 & 0.090 \\
         & top-1 cosine & 0.041 & 0.054 & 0.047 \\
         & top-5 cosine & 0.042 & 0.261 & 0.073 \\
        \midrule
        \multirow{3}{*}{Only 9 frequent labels} & LLM & 0.171 & 0.163 & 0.167 \\
         & top-1 cosine & 0.172 & 0.178 & 0.175 \\
         & top-5 cosine & 0.136 & 0.429 & 0.207 \\
        \midrule
        \multirow{3}{*}{20 coarse clusters} & LLM & 0.495 & 0.207 & 0.292 \\
         & top-1 cosine & 0.452 & 0.116 & 0.185 \\
         & top-5 cosine & 0.453 & 0.551 & 0.497 \\
    \bottomrule[1.2pt]
\multicolumn{5}{p{1.0\linewidth}}{Coarsening the label space substantially increases F1, suggesting that a large fraction of ``errors'' under strict evaluation are near misses among semantically overlapping WordNet synsets.}
\end{tabular}%
}
\label{tab:int_ablation}
\vspace{-0.5em}
\end{table}

\subsection{Ablations on Grounding Signal}
We aim to evaluate the effect of the visual cue during the instance-grounding mechanism by varying the visual cue used to direct the VLM to a target identity, systematically testing bounding boxes, single contours, and multi-contour overlays.
To isolate grounding effects from detector errors, we initialize the segmentation tracker from ground-truth person boxes and compute captioning metrics against instance-caption references.
Table~\ref{tab:grounding_ablation} shows that single-person contour achieves the best METEOR (0.328) and ROUGE (0.373), with competitive BLEU (0.420), strongly supporting our hypothesis that contours serve as a minimally invasive yet highly effective grounding primitive for precise semantic attribution.
In contrast, multi-contour overlays yield a METEOR drop to 0.192 (-0.136, 42\%) and ROUGE to 0.307 due to visual clutter confusing role attribution between color-coded instances, while single bounding boxes provide only coarse focus (BLEU 0.346, METEOR 0.248, ROUGE 0.296), underperforming the fine-grained boundary information from mask-derived contours.

\begin{table}[!t]
\centering
\caption{Instance-caption grounding ablations using ground-truth boxes for tracker initialization. Best results are in \textbf{bold}.}
\resizebox{0.95\linewidth}{!}{%
    \begin{tabular}{l|ccc}
        \toprule[1.2pt]
        \textbf{Grounding (GT init.)} & \textbf{BLEU}$\uparrow$ & \textbf{METEOR}$\uparrow$ & \textbf{ROUGE}$\uparrow$ \\
        \midrule
        Single contour (ours) & \textit{0.420} & \textbf{0.328} & \textbf{0.373} \\
        Multi-contour (color-coded) & \textbf{0.427} & 0.192 & \textit{0.307} \\
        Single bounding box & 0.346 & \textit{0.248} & 0.296 \\
    \bottomrule[1.2pt]
\multicolumn{4}{p{1.0\linewidth}}{``Single contour'' renders only the target identity; ``multi-contour'' renders multiple identities with different colors and prompts the target; ``single bounding box'' replaces contours with a target box.}
\end{tabular}%
}
\label{tab:grounding_ablation}
\vspace{-1em}
\end{table}

\subsection{Qualitative Results}
\label{subsec:qual_examples}
\begin{figure*}[t]
    \centering
    \includegraphics[width=0.95\linewidth]{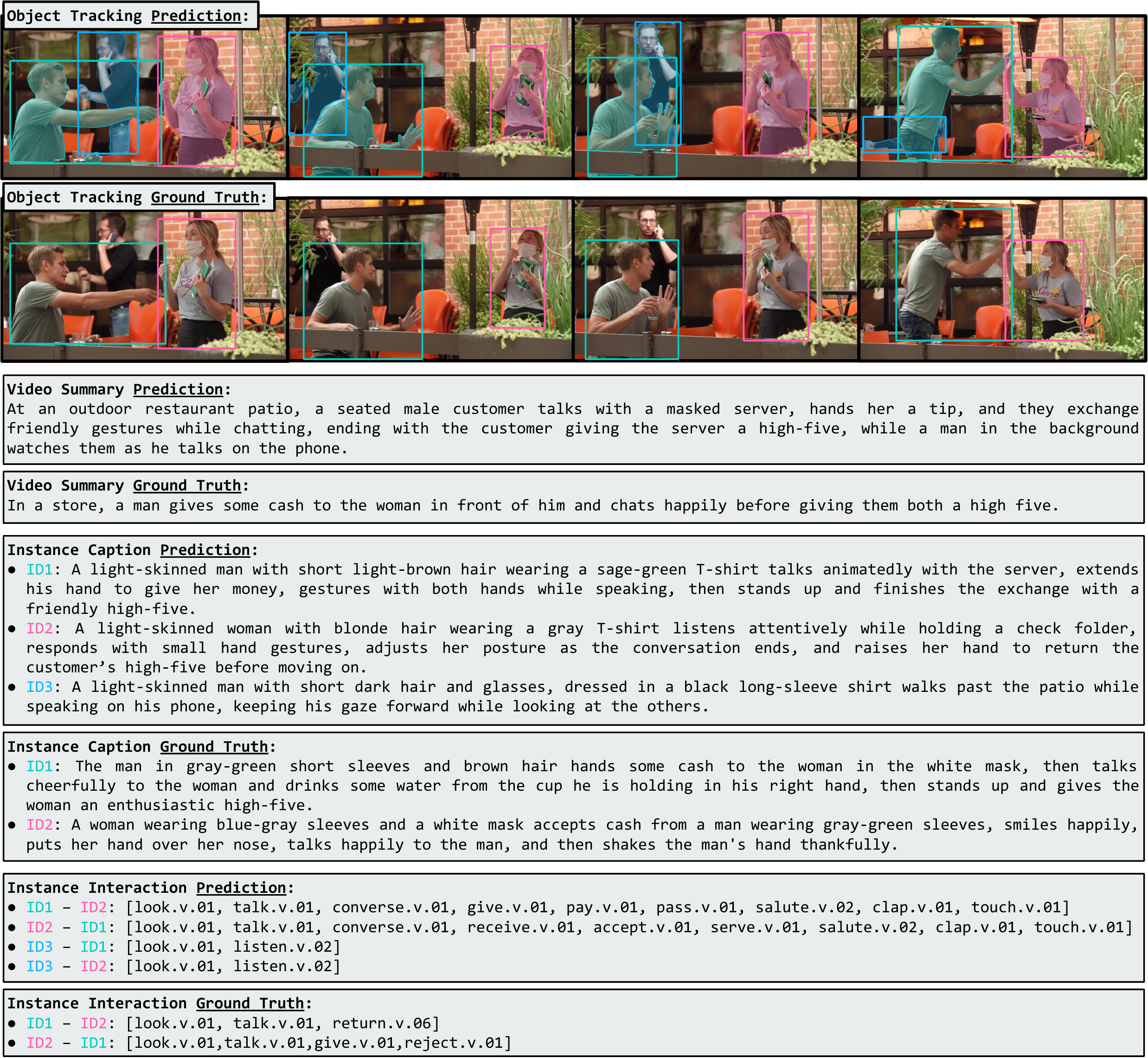}
    \caption{\textbf{\methodname semantic tracking qualitative results on BenSMOT} (sequence \texttt{jT\_8i7edEt8}). As can be seen, \methodname detects a larger number of people in the scene than the ground truth annotations and generates richer, more detailed instance captions.
    The predicted instance-level interactions are correspondingly broader and more expressive; however, they do not always exactly match the annotated ground truth interactions.
    This discrepancy leads to low exact-match interaction scores and reflects limitations of the dataset rather than purely model errors, as the selected WordNet synsets can be semantically ambiguous and the interaction annotations are not always exhaustive.}
    \label{fig:qualitative}
    \vspace{-1em}
\end{figure*}

Fig.~\ref{fig:qualitative} presents a representative qualitative example from the BenSMOT test set, comparing TF-SMOT predictions against the provided ground-truth annotations across tracking, captioning, and interaction recognition.

As shown in the top rows of Fig.~\ref{fig:qualitative}, TF-SMOT produces more fine-grained tracking outputs than the ground truth. By relying on segmentation-based tracking rather than box-level annotations, our method is able to detect and maintain identities for additional people present in the scene (\eg, background pedestrians) that are not annotated in BenSMOT.
While this reflects more complete scene understanding, these extra identities are penalized by the benchmark, which assumes a fixed and incomplete set of tracked actors. Consequently, improved recall at the tracking level can negatively impact evaluation metrics under the current protocol.

The generated video summary and instance-level captions are generally more detailed than the ground-truth language annotations.
TF-SMOT captures appearance attributes, fine-grained actions, and temporal structure (\eg, gesture sequences and transitions such as standing up or initiating a high-five) that are often simplified or omitted in the reference captions. This increased descriptiveness improves semantic faithfulness but can introduce stylistic and lexical divergence from the ground truth, which is not always rewarded by n-gram–based metrics.

Our TF-SMOT predicts a broader and more expressive set of directed interactions between tracked identities. As illustrated in Fig. 3, the model identifies multiple plausible interactions (\eg, \textit{talk}, \textit{give}, \textit{pay}, \textit{touch}, \textit{salute}) that are visually supported by the scene but not exhaustively annotated in the ground truth.
Under the strict exact-match WordNet evaluation used by BenSMOT, these additional or semantically adjacent synsets are counted as false positives, leading to low interaction scores. This behavior reflects limitations of the benchmark (specifically, ambiguous WordNet sense choices and incomplete interaction labeling) rather than a lack of semantic understanding by the model.

Overall, the qualitative results highlight a recurring pattern: TF-SMOT tends to produce more complete, fine-grained, and semantically rich outputs than the dataset annotations, but these improvements are not fully captured by the current evaluation protocol. This reinforces the need for evaluation schemes that better account for annotation sparsity, semantic overlap, and near-miss predictions in SMOT.

\section{Conclusion} \label{sec:conclusion}
We presented \methodname, a modular and training-free approach to Semantic Multi-Object Tracking (SMOT) that composes foundation models for detection, mask tracking, and video-language generation.
On the \dataset benchmark, \methodname achieves state-of-the-art performance for the \emph{tracking} component within the SMOT setting and improves video summarization and instance captioning compared to the SMOTer baseline.
At the same time, interaction recognition remains challenging under strict exact-match evaluation on the fine-grained and long-tailed WordNet label space.
We hope \methodname can serve as a strong, reproducible baseline and a diagnostic tool to motivate future work on grounded interaction understanding and more robust SMOT evaluation protocols.
We encourage future SMOT work to report complementary views of interaction quality that disambiguate ontology issues from genuine semantic failures.
One simple option is to provide head/tail splits (frequent vs. rare interactions) and to add macro-averaged metrics that reduce dominance of frequent classes.
Furthermore, SMOT would benefit from semantic proximity metrics, e.g., counting near-miss synsets as partially correct based on WordNet similarity or within-cluster correctness.
Finally, because BenSMOT interactions are directed, reporting direction-only accuracy (subject/object correctness) can isolate whether errors come from role assignment or from verb/sense selection.

\noindent \textbf{Potential negative societal impact.}
Semantic multi-object tracking combines identity-aware tracking with language outputs such as captions and interaction labels.
As with other tracking and video analytics technologies, this can raise societal concerns related to privacy, surveillance, and misuse in sensitive contexts (e.g., public-space monitoring or workplace oversight) without informed consent.
Moreover, erroneous interaction predictions could lead to incorrect inferences about people’s intentions or behaviors, with disproportionate harms in high-stakes settings.
Responsible deployment should include clear governance, safeguards against misuse, and careful evaluation of failure modes and biases.
In addition, composing large pretrained models may inherit societal biases present in web-scale data.
For instance, generated descriptions may overemphasize demographic attributes or hallucinate actions that are not visually supported.
To mitigate these risks, practitioners should avoid using semantic tracking as the sole basis for decisions, provide uncertainty-aware interfaces, and evaluate performance across subpopulations and contexts when such data is available.

\section*{Acknowledgments}
\small We acknowledge EuroHPC Joint Undertaking for awarding us access to MareNostrum5 at BSC, Spain. This work was supported by the EU Horizon ELIAS (No. 101120237), ELLIOT (No. 101214398), TURING (No. 101215032), IAMI (No. 101168272), and PATTERN (No. 101159751) projects. This work was carried out in the Vision and Learning joint laboratory of FBK and UniTN.

{
    \small
    \bibliographystyle{ieee}
    \bibliography{main}
}

\clearpage
\setcounter{page}{1}
\maketitlesupplementary

\section{Prompts}
\label{sec:supp_prompt}
We provide the exact prompts used by our \methodname for video summarization, instance caption, verb extraction, and the selection of WordNet synsets.

\begin{lstlisting}[caption={InternVideo2.5 prompt for video summaries.},label={tab:prompt_video_summary}]
You are a precise video-understanding assistant.  
When given a video, produce **one concise sentence** that states:

**Where** the action happens (room type, environment, etc.)  
**Who** the visible main actors are (woman, two boys, etc.)  
**What** they are doing (main action, key objects)  

Use the present tense, third-person voice, and end with a period.  
Avoid extra words; aim for less than 25 tokens.

### Examples
Video 1 -> Caption: In a room, a woman brushes the teeth of a baby who is crying.  
Video 2 -> Caption: In the restroom, a man squeezes toothpaste and brushes the teeth of the boy beside him.  
   [...]

### Task
Attached video: Provide the caption.
\end{lstlisting}

\begin{lstlisting}[caption={InternVideo2.5 prompt for instance captions.},label={tab:prompt_instance_caption}]
Generate a detailed, single-sentence caption that focuses exclusively on the individual outlined by a {color} contour in the video scene, without mentioning the contour itself. 

The caption should: 
    - Appearance: Objectively detail the person's physical features, including clothing (with specific colors, patterns, styles) and any distinctive accessories. 
    - Actions & Expressions: Describe their body posture, gestures, facial expressions, and dynamic movements (e.g., brushing teeth, talking, or interacting with objects) in a sequential and natural manner. 
    - Interactions: Seamlessly incorporate any relevant interactions with objects or other people into a coherent, story-like narrative that remains focused on the {color}-contoured individual. 
    - Style Reference: Your caption should be objective and richly descriptive, mirroring the narrative style of these examples: 'A woman wearing a red headscarf is watching a girl wearing a blue short-sleeve brushing her teeth while talking and making a brushing gesture with her right hand to teach the girl to speak.', 'A man in blue and white floral shirt and red and gray hat holding a cup in his right hand is talking to man in dark gray cotton suit, he then takes a package in his left hand from the man in the dark gray cotton suit, switches the cup to his left hand, and shakes his hand.'. Ensure the output is coherent, precise, and fully captures the dynamic context of the scene.
\end{lstlisting}

\begin{lstlisting}[caption={LLaMA prompt for directed interaction verb extraction.},label={tab:prompt_verb_extraction}]
You are a state-of-the-art Large Language Model whose sole task is to extract **human-to-human interaction** verbs from scene descriptions.     
## TASK
Given a JSON of individual behavior descriptions, identify every directed interaction between people.
## DEFINITIONS
- **Subject** = the actor initiating the interaction.  
- **Object** = the receiver or target of the interaction.  
- **Interaction** = a single verb in **present tense**, denoting what the Subject does to or with the Object (e.g., `"look"`, `"give"`, `"converse"`).
## INPUT
A single JSON object where:
- **Keys** are person IDs (e.g. `"ID_1"`, `"ID_2"`, ...). 
- **Values** are natural-language scene descriptions mentioning behaviors, gestures, exchanges, motion relative to other people.

Example:
```json
{
   [...]
}
```
## OUTPUT
Return **only** a JSON object (no extra text) mapping each `subject_ID` to a dictionary of `object_ID: [interactions]`.

Rules:
- **Exact IDs**: Use the same IDs from the input.
- **Verb List**: One verb per interaction. Present-tense, infinitive base form (no "-ing", no tense variants).
- **Empty Lists**: If a subject performs no observable actions on an object, include the key with an empty list.
- **Self-interactions**: Omit or return an empty list for `subject == object`.

Example output for the above input:

```json
{
   [...]
}
```
**Important**: Do **not** include any additional keys or commentary-only the final JSON.
\end{lstlisting}

\begin{lstlisting}[caption={Prompt for WordNet ID selection from top-$K$ candidates.},label={tab:prompt_wordnet_selection}]
Task: You will receive one target sentence and a list of candidate WordNet definitions.
    Select the single definition whose meaning best matches the sentence.
    Sentence:{sentences[obj_id]} 
    Definitions (same order as supplied): <number>|<wordnet-id>|<definition-text>
    Response format: Return only a JSON object with the chosen ID, no extra text. {"wordnet-id": "<number>"}(Replace <number> with the identifier from the selected definition.)
\end{lstlisting}

\end{document}